\documentclass[conference]{IEEEtran}
\IEEEoverridecommandlockouts
\usepackage{cite}
\usepackage{amsmath,amssymb,amsfonts}
\usepackage{algorithmic}
\usepackage{graphicx}
\usepackage{textcomp}
\usepackage{xcolor}
\def\BibTeX{{\rm B\kern-.05em{\sc i\kern-.025em b}\kern-.08em
    T\kern-.1667em\lower.7ex\hbox{E}\kern-.125emX}}
\long\def\comment#1{}

\makeatletter
\newcommand{\linebreakand}{%
  \end{@IEEEauthorhalign}
  \hfill\mbox{}\par
  \mbox{}\hfill\begin{@IEEEauthorhalign}
}
\makeatother

\title{Time-Window Group-Correlation Support vs. Individual Features: A Detection of Abnormal Users}

\author{
\IEEEauthorblockN{Lun-Pin Yuan}
\IEEEauthorblockA{Penn State University \\
lunpin@psu.edu}
\and
\IEEEauthorblockN{Euijin Choo}
\IEEEauthorblockA{Qatar Computing Research Institute \\
echoo@hbku.edu.qa}
\and
\IEEEauthorblockN{Ting Yu}
\IEEEauthorblockA{Qatar Computing Research Institute \\
tyu@hbku.edu.qa}
\linebreakand 
\IEEEauthorblockN{Issa Khalil}
\IEEEauthorblockA{Qatar Computing Research Institute \\
ikhalil@hbku.edu.qa}
\and
\IEEEauthorblockN{Sencun Zhu}
\IEEEauthorblockA{Penn State University \\
sxz16@psu.edu}
}

\usepackage{amsmath}
\usepackage{amssymb}
\usepackage{algorithmic}
\usepackage[ruled,vlined,linesnumbered]{algorithm2e}
\usepackage{soul}
\usepackage{subfigure}
\usepackage{todonotes}

\SetKwInput{Input}{Input}
\SetKwInput{Output}{Output}

\newcommand{\nop}[1]{}

\newcommand{\behfig}[3]{
	{\begin{figure}
	\centering
	\includegraphics[width=#2\linewidth]{behfigures/#1}
	\caption{#3}
	\label{behfig:#1}
	\end{figure}}
}

\newcommand{\behfigw}[3]{
	{\begin{figure*}
	\centering
	\includegraphics[width=#2\textwidth]{behfigures/#1}
	\caption{#3}
	\label{behfig:#1}
	\end{figure*}}
}

\begin{document}

\maketitle
\thispagestyle{plain}
\pagestyle{plain}

\begin{abstract}
Autoencoder-based anomaly detection methods have been used in identifying anomalous users from large-scale enterprise logs with the assumption that adversarial activities do not follow past habitual patterns. Most existing approaches typically build models by reconstructing single-day and individual-user behaviors. However, without capturing long-term signals and group-correlation signals, the models cannot identify low-signal yet long-lasting threats, and will wrongly report many normal users as anomalies on busy days, which, in turn, lead to high false positive rate. In this paper, we propose \textit{ACOBE}, an \textit{A}nomaly detection method based on \textit{CO}mpound \textit{BE}havior, which takes into consideration long-term patterns and group behaviors. \textit{ACOBE} leverages a novel behavior representation and an ensemble of deep autoencoders and produces an ordered investigation list. Our evaluation shows that \textit{ACOBE} outperforms prior work by a large margin in terms of precision and recall, and our case study demonstrates that \textit{ACOBE} is applicable in practice for cyberattack detection.

\end{abstract}

    {\section{Introduction}

Emerging cyber threats such as data breaches, data exfiltration, botnets, and ransomware have caused serious concerns in the security of enterprise infrastructures~\cite{cyberthreat1, cyberthreat2}. 
The root cause of a cyber threat could be a disgruntled insider or a newly-developed malware.  What is worse, emerging cyber threats are more difficult to be identified by signature-based detection methods, because more and more evasive techniques are available to adversaries.  To identify the emerging cyber threats before they can cause greater damage, anomaly detection upon user behaviors has attracted focuses from large-scale enterprises~\cite{svddc, nlsalog16, lifelong26, liuliu1, liuliu2, pca01, pca02}, as anomaly detection enables security analysts to find suspicious activities that could be aftermath of cyber threats (including cyberattacks and insider threats).  Adversarial activities often manifest themselves in abnormal behavioral changes compared to past habitual patterns.  Our goal is to find such abnormal behavioral changes by learning past habitual patterns from organizational audit logs.

Autoencoders are one of the most well-known anomaly detection techniques~\cite{deepsurvey1, deepsurvey2, deepsurvey3}.  They are attractive to security analysts because of their robustness to domain knowledge.  Briefly speaking, an autoencoder-based anomaly detection model only learns how to reconstruct normal data; hence, in events of poor reconstruction, the input data is highly likely to be abnormal.  Detailed domain knowledge is no longer required, because the complex correlation among different activities is captured by learning the normal distribution from normal activities.  Furthermore, autoencoder learns features from unlabeled data in an unsupervised manner, where human interference is reduced. 


However, similar to other typical anomaly detection methodologies, the autoencoder-based approaches also suffer from the overwhelming number of false positive cases.  The challenge is that, while an anomaly detection model is required to be sensitive to abnormal events in order to raise anomaly alerts, the model can also be so sensitive to normal behavioral deviation that it often wrongly reports a normal deviation as an anomaly (hence a false positive).  
This challenge leads to the following question: \textit{how to further differentiate abnormal events from normal events}.  While prior approaches work with input data that meets the data quality requirements (e.g., completeness, availability, consistency) for cybersecurity applications~\cite{dataquality}, important factors such as misconduct timeliness and institutional environment are not considered.
To reduce false positives, we argue that it is important to also examine \textit{long-term} signals and \textit{group-correlation} signals, as opposed to typical autoencoder-based approaches that only examine \textit{single-day} and \textit{individual-user} signals. 

The limitation of only examining \textit{single-day} and \textit{individual-user} signals is twofold.
First, without capturing \textit{long-term} signals, a model cannot identify low-signal yet long-lasting threats.  Certain cyber-threat scenarios do not cause immediate behavioral deviation, but progressively cause small yet long-lasting behavioral deviation; for example, an insider threat is a scenario where a disgruntled employee stealthily leaks sensitive data piece-by-piece over time~\cite{certdataset2,certdataset1}.  What is worse, without \textit{long-term} signals, a model will likely wrongly reports many normal users as anomalies on busy days (e.g., working Mondays after holidays) due to the massive burst of events in a short term.  This further leads to the second limitation: without capturing \textit{group-correlation} signals, a model cannot reduce its false positive rate in occasions (e.g., environmental change) where many users have common burst of events.  For example, common traffic bursts occur among many users when there is a new service or a service outage.  By examining behavioral \textit{group-correlation}, a model may figure out that the common behavioral burst is indeed normal, assuming the more behavioral correlation a user has with the group, the less likely the user is abnormal.

To address these fundamental limitations, we propose a different methodology, which involves a novel behavioral representation.  We refer to this representation as a \textit{compound behavioral deviation matrix}.  Each matrix encloses individual behaviors and group behaviors across multiple days. Having compound behaviors, we then apply anomaly detection models, which are implemented with an ensemble of deep fully-connected autoencoders.  We propose \textit{\textbf{ACOBE}}, an \textit{\textbf{A}nomaly detection method based on \textbf{CO}mpound \textbf{BE}havior}.  ACOBE has three steps in its workflow (Figure~\ref{behfig:workflow.pdf}): it first derives compound behavioral deviation matrices from organizational audit logs; then, for each user, it calculates anomaly scores in different behavioral aspects using an ensemble of autoencoders; lastly, having anomaly scores, ACOBE produces an ordered list of the most anomalous users that need further investigation.  In summary, we make the following contributions.


\begin{enumerate}
\item We propose a novel behavioral representation which we refer to as the \textit{compound behavioral deviation matrix}.  It profiles individual and group behavior over a time window (e.g., several days).  We further apply deep fully-connected autoencoder upon such representation in order to find anomalous users. Our model outputs an ordered list of anomalous users that need to be investigated.

\item We evaluate ACOBE upon a synthesized and labeled dataset which illustrates insider threats.  Our results show that ACOBE outperforms our re-implementation of a prior work.  With four abnormal users out of 929 users, ACOBE achieves $99.99\%$ AUC (the area under the ROC curve).  ACOBE effectively puts abnormal users on top of the investigation list ahead of normal users.  
\item   We demonstrate with case studies that ACOBE can be applied in practice for realistic cyber threats, including botnets and ransomware attacks.  
\end{enumerate}


}

    {\section{Related Work}\label{behsec:relatedwork}

Most anomaly detection models are zero-positive machine learning models that are trained by only normal (i.e., negative) data.  These models are then used in testing whether an observation is normal or abnormal, assuming unforeseen anomalies do not follow the learned patterns.  Kanaza et al.~\cite{svddc} integrated supports vector data description and clustering algorithms, and Liu et al.~\cite{nlsalog16} integrated K-prototype clustering and k-NN classification algorithms to detect anomalous data points, assuming anomalies are rare or accidental events.  When prior domain knowledge is available for linking causal or dependency relations among subjects, objects and operations, graph-based anomaly detection methods (such as Elicit~\cite{pca02}, Log2Vec~\cite{log2vec}, Oprea et al.~\cite{nlsalog20}) could be powerful.  When little prior domain knowledge is available, Principal Component Analysis (PCA) based anomaly detection methods (for example, Hu et al.~\cite{pca01} proposed an anomaly detection model for heterogeneous logs using singular value decomposition) could be powerful.  Contrary to zero-positive anomaly detection are semi-supervised or online learning anomaly detection, in which some anomalies will be available over time for training the model~\cite{lifelong}.

The autoencoder framework is a PCA approach that is widely used in anomaly detection.  A typical autoencoder-based anomaly detection method learns how to reconstruct normal data.  It then detects anomalies by checking whether the reconstruction error of an observation has exceeded a threshold.  To detect anomalies, Zong et al. proposed deep autoencoding Gaussian mixture models~\cite{lifelong48} and Chiba et al. proposed autoencoders with back propagation~\cite{chiba}. Sakurada and Yairi proposed autoencoders with nonlinear dimensionality reduction~\cite{lifelong31}. Lu et al. proposed an autoencoder constrained by embedding manifold learning, MC-AEN~\cite{exploitembed}. Nguyen et al. proposed a variational autoencoder with gradient-based anomaly explanation, GEE~\cite{gee}. Wang et al. proposed a self-adversarial variational autoencoder with Gaussian anomaly prior assumption, adVAE~\cite{advae}. Alam et al. proposed an ensemble of autoencoders accompanied by K-mean clustering algorithm, AutoPerf~\cite{lifelong15}. Mirsky et al. proposed an ensemble of lightweight autoencoders, Kitsune~\cite{lifelong26}. Liu et al. proposed an ensemble of autoencoders for multi-sourced heterogeneous logs~\cite{liuliu1, liuliu2}. Chalapathy et al.~\cite{rcae} and Zhou et al.~\cite{lifelong47} proposed robust autoencoders.  For other anomaly detection methods, detail surveys can be found in~\cite{deepsurvey1, deepsurvey2, deepsurvey3, deepsurvey4, survey1, nlsalog10, nlsalog25}.


Each of the above autoencoder work focuses on either the optimization of a particular learning algorithm without cybersecurity context~\cite{lifelong48,lifelong31,exploitembed,advae,lifelong15,rcae,lifelong47}, or the optimization of a learning framework with very specific cybersecurity context (i.e., network intrusion detection system)~\cite{chiba,gee,lifelong26}.  
However, though they provide fruitful insights in their particular domains, it is hard to apply their model or framework to the other anomaly detection problems (including ours---\textit{\textbf{detection of abnormal users in a large-scale organization}}) due to the requirement difference for the input data.  
For example, with statistical network-traffic features for individual transactions, it is difficult to discover a disgruntled insider who exfiltrates proprietary secrets piece-by-piece in a long run.

Regardless of model and framework design, any input data that does not meet the requirement may fail the detection methodology, and this is commonly known as the \textit{data quality challenge}~\cite{dataquality}.
While Sundararajan et al.~\cite{dataquality} discussed six data-quality requirements (i.e., completeness, measurement accuracy, attribute accuracy, plausibility and availability, origination, and logical consistency) for cybersecurity applications, important factors such as misconduct timeliness and institutional environment are not considered.
Hence, in this paper, we utilize two requirements---\textit{long-term signals} and \textit{group-behavior} should be examined---to address the anomaly detection problem challenges and limitations of the existing work (e.g., high false-positive rate).
We shall further show the benefits of our data-quality requirement in Section~\ref{behsec:evaluation1} and~\ref{behsec:evaluation2}.
}

    {\section{Motivation}\label{behsec:motivation}

Anomaly detection upon user behaviors enables security analysts to find suspicious activities caused by cyber threats including insider threats (Section~\ref{behsec:evaluation1}) and cyber attacks (Section~\ref{behsec:evaluation2}). Typical anomaly detection methods often suffer from the overwhelming number of false positives due to the sensitivity to normal behavioral deviation. Moreover, such methods often only output anomaly labels (i.e., either \textit{normal} or \textit{abnormal}). However, if a model provides only anomaly labels, security analysts will be overwhelmed by heavy workload of investigation. Hence, in practice, it is more preferable to have an ordered list of users that need to be investigated~\cite{oprea2018made}. We thus define an anomaly detection problem as follows: \textit{\textbf{given a set of per-user activities, provide an investigation list of usernames that need to be orderly investigated}}.  On the top of such a list are the most abnormal users.

Autoencoders have been widely used in solving such an anomaly detection problem, as it is capable of learning \textit{what are normal} in order to find \textit{what are abnormal}.  Among the aforementioned anomaly detection methods, we find Liu et al.~\cite{liuliu1} and Hu et al.~\cite{pca01}'s works are representative.  They are similar anomaly detection methods that reconstruct single-day individual-user behaviors.  However, single-day user-behavior reconstruction is not an ideal solution for identifying cyber threats.  We argue that it is important to examine \textit{long-term} signals and \textit{group-correlation} signals to reduce the false-positive rate, because of the following reasons.

First, certain cyber compromise scenarios do not cause immediate behavioral deviation, but progressively cause small long-lasting behavioral deviation in different behavioral aspects across multiple days.  Take Zeus botnet malware as an example, once triggered, it modifies registry values (deviation in system configuration). After a few days, it communicates with the C\&C server and acts maliciously (deviation in network traffic).  Since these two behavioral deviations occur on different days, only models that examine long-term behaviors can identify such an anomaly.  In contrast, single-day user-behavior reconstruction 
may fail to identify the threats, 
and it may also wrongly give high anomaly scores to busy days (e.g., working Mondays and make-up days). 

The granularity of feature measurements is also an important factor in building profiles to accurately capture normal user behavior. Concretely, users often tend to have more human-initiated activities (e.g., sending emails, browsing web pages, writing documents) during working hours, but more computer-initiated activities (e.g., system updates, system backups, network retries due to no password input) during off hours. Therefore,  our approach also captures behaviors over multiple time-frames (e.g., \textit{working hours} and \textit{off hours}) within each day of the measurement window.

Second, there often exists certain behavioral correlation between an individual user and its group due to environmental change or locale change.  Take environmental change as an example, when there is a new service or service outage, one can expect correlated unrecognized traffics or correlated retry traffics, respectively.  Take locale change as another example, one can expect more human-initiated activities during working hours on working days, as opposed to during off hours or holidays.  Based on this observation, we make the following hypothesis: the greater behavioral correlation a user has with the group, the less likely the user is compromised.  A model without incorporating the group behavior may not only be ineffective in identifying anomalies, but also wrongly give high anomaly scores to individual users in events of environmental or locale changes.  Hence, we propose to incorporate both individual-user behaviors and group behaviors to better capture normal behavior and avoid obvious false positives that may result from unusual yet common activities of users.

}

    {\section{Our Methodology}\label{behsec:design}

\behfigw{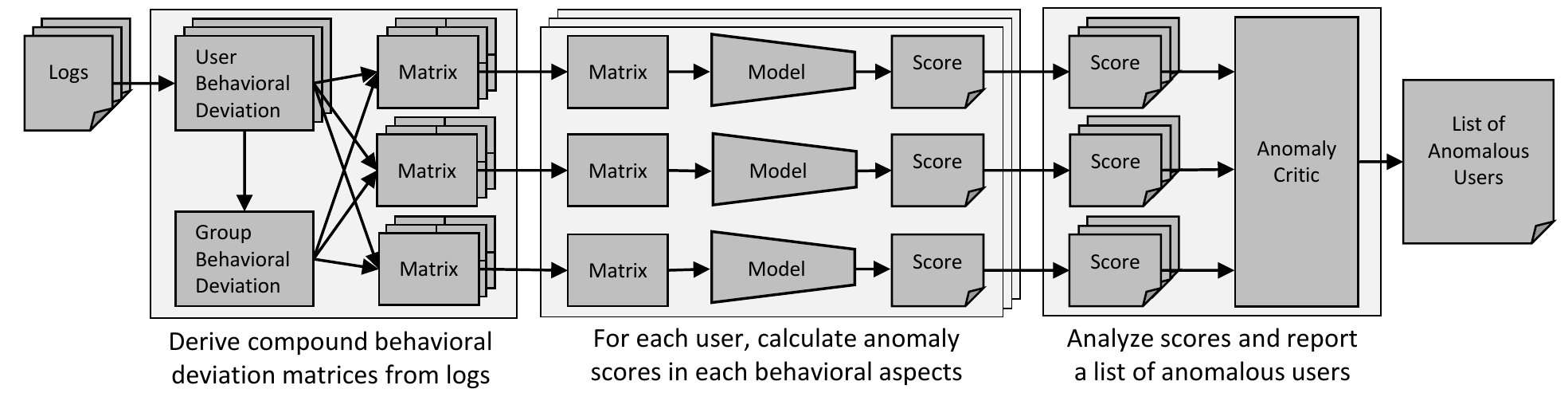}{1}{ACOBE Workflow}

To address the challenges discussed in the previous subsection, we present \textbf{\textit{ACOBE}}, an \textit{\textbf{A}nomaly detection method based on \textbf{CO}mpound \textbf{BE}havior}, where a compound behavior encloses individual-user behaviors and group behaviors across multiple time-frames and a time window in days.  Having compound behavior, we then apply anomaly detection implemented with deep fully-connected autoencoders.  Figure~\ref{behfig:workflow.pdf} illustrates the workflow: (1) ACOBE first derives compound behavioral deviation matrices from organizational audit logs, (2) for each user, ACOBE then calculates anomaly scores in different behavioral aspects using an ensemble of autoencoders, and (3) having anomaly scores, ACOBE finally produces an ordered list of users that need further investigation.

\subsection{Compound Behavioral Deviation Matrix}

\behfig{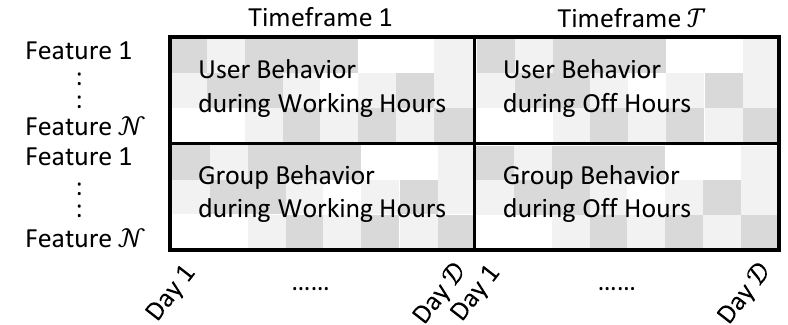}{1}{Compound Behavioral Deviation Matrix}

A compound behavioral deviation matrix encloses deviation of individual user behavior and group behavior across multiple time-frames and multi-day time window.  Figure~\ref{behfig:heatmap.pdf} illustrates an example of a compound behavioral deviation matrix with $\mathcal{F}$ features, $\mathcal{D}$ days, and $\mathcal{T}=2$ time-frames (i.e., working hours 6am-6pm and off hours 6pm-6am).  Each feature in user behavior represents a normalized characteristic of an aggregated behavior, including but not limited to, the numbers of successful logons, file accesses, failure HTTP queries (during the specific time-frame on specific day indicated by columns).  Since feature selection is domain-specific, we leave the details of features in the evaluation (Section~\ref{behsec:evaluation1}) and case-study sections (Section~\ref{behsec:evaluation2}).  Features in group behavior are derived by averaging the corresponding features of all users in the group.  Note that, how these four components are stacked together is not important (alternative stackings are applicable), because matrices will be flattened before going through the anomaly detection models.  

We derive our deviation measurement $\sigma_{f,t,d}$ for feature $f$ in time-frame $t$ on day $d$ with the below equations. $m_{f,t,d}$ denotes the numeric measurements (of feature $f$ in time-frame $t$ on day $d$). $\vec{h}_{f,t,d}$ denotes the vector of history numeric measurements (in $\omega-1$ days before day $d$, where $\omega$ is the window size in days). $std (\vec{h}_{f,t,d})$ denotes the standard deviation of history measurements ($std$ is set to $\epsilon$ if it is less than $\epsilon$ to avoid divide-by-zero exception). $\delta_{f,t,d}$ denotes the variance of numeric measurement and $\sigma_{f,t,d}$ denotes the final behavioral deviation which is bounded by $\Delta$.  We bound $\sigma_{f,t,d}$ by a large $\Delta$, as it is equivalently anomalous when $|\delta_{f,t,d}| \geq \Delta$. For example, variances larger than $\Delta=3$ are equivalently \textit{very abnormal}, assuming the numeric measurements follow Gaussian distribution.  Note that, in events when users slowly shift their normal behavioral patterns over time, their compound behavioral deviation matrices will not show increasing deviation over time, as the history $\vec{h}_{f,t,d}$ (from which deviations are derived) slides through time and will always cover the recent shift.

\begin{align*}
m_{f,t,d} = & \text{ numeric measurements of feature } f \\
& \text{ in timeframe } t \text{ on day } d \\
\vec{h}_{f,t,d} = & [ m_{f,t,i} | i: d-\omega+1 \leq i < d ] \\
std (\vec{h}_{f,t,d}) = & 
\begin{cases}
    \epsilon, \text{ if } \text{standard-deviation } (\vec{h}_{f,t,d}) < \epsilon \\
    \text{standard-deviation } (\vec{h}_{f,t,d}), \text{ otherwise}
\end{cases} \\
\delta_{f,t,d} = & \frac{m_{f,t,d} - \text{mean } (\vec{h}_{f,t,d})}{\text{std } (\vec{h}_{f,t,d})} \\
\sigma_{f,t,d} = & 
\begin{cases}
    \Delta, & \text{if } \delta_{f,t,d} > \Delta \\
    -\Delta, & \text{if } \delta_{f,t,d} < -\Delta \\
    \delta_{f,t,d}, & \text{otherwise}
\end{cases}
\end{align*}

Weights are applied to features, as different behaviors have different importance in capturing user behavior; for example, frequent and chaotic \textit{file-read} activities are often less critical than rarer \textit{file-write} activities.  Since the anomaly scores are essentially the reconstruction errors of features, applying weights to features can scale-down unimportant features and thus the partial errors introduced by unimportant features. Consequently, it makes ACOBE to focus only on reconstructing important features while being more resilient to noise introduced by unimportant features.  To automatically reflect the relative importance without conducting empirical studies upon individual users, Hu et al.~\cite{pca01} applied weights to features based on Term Frequency-Inverse Document Frequency (TF-IDF) measurements, which was originally designed for measuring the amount of information a particular text subject provides based on text frequency.  Similarly, ACOB  offers the option of introducing the following feature weights $w_{f,t,d}$.  

\begin{align}\label{beheq:weight}
w_{f,t,d} &= \frac{1}{\log_2 \Bigg( \text{ max } \Big (std (\vec{h}_{f,t,d}), 2 \Big) \Bigg) }
\end{align}

The equation is based on the \textit{log-normalized TF weight}, which is defined by $\textit{TF}_x = \log (1+\text{frequency of a term } x)$; that is, the less the frequency, the less information a term $x$ could provide.  Yet, unlike terms, the higher $std$ (or equivalently more chaotic), the less information a feature $f$ could provide.  To serve our need, we inverse the equation $\textit{TF}_x$ and substitute standard deviation for frequency, so that the weights are lower for chaotic features but higher for consistent features by design. However, it cannot be infinitely high for constantly static features with very small $\text{std }(\vec{h}_{f,t,d})$, or otherwise ACOBE would be overly sensitive to small changes of static features.
Therefore, a minimum value of two is given to the logarithm function, and we change the base to two so that the maximum value of weights is bounded to one. In other words, activities with small standard deviation less than two shall have equal weights of value one.  


\subsection{Anomalous Deviation Detection Model}

\behfig{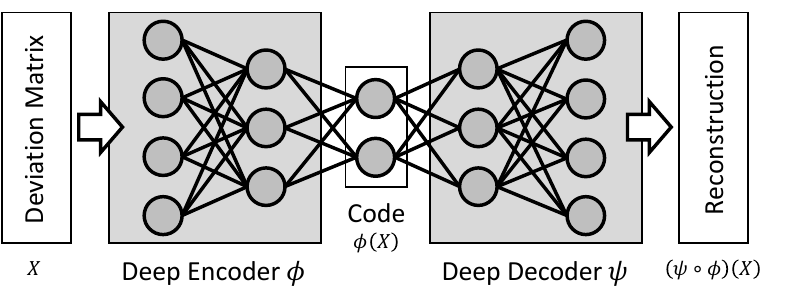}{1}{Deep Fully-Connected Autoencoder}

As shown in Figure~\ref{behfig:workflow.pdf}, we leverage an ensemble of autoencoders, each of which identifies behavioral anomalies in terms of a designated \textit{\textbf{behavioral aspect}}, where a behavioral aspect is a set of relevant behavioral features; for example, (1) \textit{file-access} aspect includes \textit{file-read}, \textit{file-write}, and \textit{file-delete} activities, (2) \textit{network-access} aspect includes \textit{visit}, \textit{download}, and \textit{upload} activities, and (3) \textit{configuration} aspect includes \textit{registry-modification}, \textit{password-modification}, and \textit{group-modification}.  

We leverage fully-connected autoencoders in identifying anomalous compound behavioral deviation matrices.  Trained with only normal matrices, an autoencoder is able to reconstruct only normal matrices with minimal reconstruction errors.  In events of high reconstruction errors, the abnormal matrices are likely aftermath of compromised users or unusual legitimate user activity.  Figure~\ref{behfig:autoencoder.pdf} illustrates an example of a deep fully-connected autoencoder.  It is trained by minimizing $\phi, \psi = \text{arg min}_{\phi, \psi} \|X - (\psi \circ \phi) (X) \| $, where $\phi$ is a multi-layer fully-connected encoder, $\psi$ is a multi-layer fully-connected decoder, $X$ is the input matrix, and $(\psi \circ \phi) (X)$ is the reconstructed matrix.  In between, the code $\phi(X)$ essentially encloses the characteristics of the input matrix.  Vincent et al.~\cite{stackautoencoder} suggested that stacking layers can help with reducing reconstruction error between the input and the output.  


\subsection{Anomaly Detection Critic}

After retrieving anomaly scores (which essentially are reconstruction errors) from an ensemble of autoencoders, our anomaly detection critic then produces a list of users that need to be orderly investigated.  
Recall that, we do not assign anomaly labels (e.g., \textit{normal} or \textit{abnormal}) to users, because providing labels without ordering is not really helpful in the context of anomaly detection, which typically has overwhelming numbers of false-positive cases.  
It is more preferable to have an ordered list of users that need to be investigated~\cite{oprea2018made}.

The investigation priority of a user is derived based on the $N$-th highest rank of the user in different behavioral aspects (that is, in more aspects is a user top anomalous, the more anomalous the user is).  For example, say $N=2$ and a user is ranked at 3rd, 5th, 4th in terms of in-total three behavioral aspects, since 4th is the 2nd highest rank of this user, this user has a investigation priority of 4.  The investigation list is sorted based on these priorities (the smaller the number, the higher the priority); if the 4 is the highest priority in the list, then this example user will be put on top of the list.  
A simpler way to understand $N$ is to imagine that $N$ is the number of votes required from each behavioral aspects.
Having the investigation list, security analysts may decide how many users they want to investigate.  For example, they can investigate only top $1\%$ of the users, or stop even earlier if they have checked a certain number of users and found nothing suspicious.  


\begin{algorithm}
\DontPrintSemicolon
\SetAlgoLined
\caption{Anomaly Detection Critic}
\label{behalg:critic}
\Input{ the number $N$, and a set of users $\mathbb{U} = \{ \mathcal{U}_1,\mathcal{U}_2, \dots \}$, where each user has ranks $\mathcal{U}_i.\mathbb{R}=\{\mathcal{R}_1, \mathcal{R}_2, \dots\}$ in different aspects } 
\Output{ a list of users ordered by priority} 
$\mathbb{P} \gets$ a list that stores $(user, priority)$ tuples \;
\ForEach{$\mathcal{U}_i \in \mathbb{U}$}{
    $ranks \gets $ sort $\mathcal{U}_i.\mathbb{R}$ \;
    $priority \gets ranks [N-1]$ \textit{// index starts from 0}\;
    append tuple $(\mathcal{U}_i, priority)$ into $\mathbb{P}$ \;
}
$list \gets $ sort $\mathbb{P}$ based on priority \;
\Return $list$ \;
\end{algorithm}



}

    {\section{Evaluation Upon Synthesized Data}\label{behsec:evaluation1}

We evaluate our proposed work upon the CERT Division Insider Threat Test Dataset~\cite{certdataset1, certdataset2}.  It is a synthesized dataset that simulates a large-scale organizational internet.  

\textbf{Implementation: } We implement the autoencoder model with Tensorflow 2.0.0.  Each fully connected layer is implemented with \textit{tensorflow.keras. layers.Dense} activated by ReLU.  The numbers of hidden units at each layer in the encoder are 512, 256, 128, and 64; the numbers of hidden units in the decoder are 64, 128, 256, and 512.  Between layers, Batch Normalization~\cite{batchnormalization} is implemented with \textit{tensorflow.keras.layers.Batch-Normalization}; batch normalization serves the purpose of optimizing training procedure. To train the model, Adadelta optimizer is used in minimizing Mean-Squared-Error (MSE) loss function.  Before feeding compound behavioral deviation matrices, we flatten the matrices into vectors, and transform the deviations from close-interval $[-\Delta, \Delta]$ to $[0, 1]$.
\begin{align*}
\text{MSE} = \frac{1}{n} \sum_{i=1}^n \Big( X_i - (\psi \circ \phi) (X_i) \Big)^2 
\end{align*}

\subsection{Dataset Description and Pre-processing}

\subsubsection{\textbf{Insider Threat Scenarios}}
The dataset provides anomaly labels for five pre-defined threat scenarios. Among them, however, we evaluate our approach for the below scenarios, as we are particularly interested in user-based anomaly detection.  

\begin{enumerate}
\item User who did not previously use removable drives or work after hours begins logging in after hours, using a removable drive, and uploading data to Wikileaks.org. Leaves the organization shortly thereafter.
\item User begins surfing job websites and soliciting employment from a competitor. Before leaving the company, they use a thumb drive (at markedly higher rates than their previous activity) to steal data.
\end{enumerate}

\subsubsection{\textbf{Training Sets and Testing Sets}}

The dataset has two sub-datasets, namely, r6.1 and r6.2.  Both r6.1 and r6.2 span from 2010-01-02 to 2011-05-31. Each subset contains one instance of each threat scenario; hence, there are four abnormal users in the four corresponding groups.  We define their groups by their organizational departments (i.e., the third-tier organizational unit) listed in the LDAP logs.  There are in total 925 normal users in these four groups.  Since the four threat scenarios occur in different times, we select the training sets and the testing sets for each scenario accordingly; yet, the detection metrics in terms of true positives (TPs), false positives (FPs), and false negatives (FNs) are put together into derivation of $F_1$ scores.  For each scenario, the training set includes the data from the first collection day until roughly one month before the date of the labeled anomalies, and the testing set includes the dates from then until roughly one month after the labeled anomalies.  Take r6.1 Scenario 2 as example, since the anomalies span from 2011-01-07 to 2011-03-07, we build the training set from 2010-01-02 to 2010-11-30, and the testing set from 2010-12-01 to 2011-03-30.  The window size ($\omega$) is set to 30 days.  

\subsubsection{\textbf{Behavioral Feature Extraction}}

\behfig{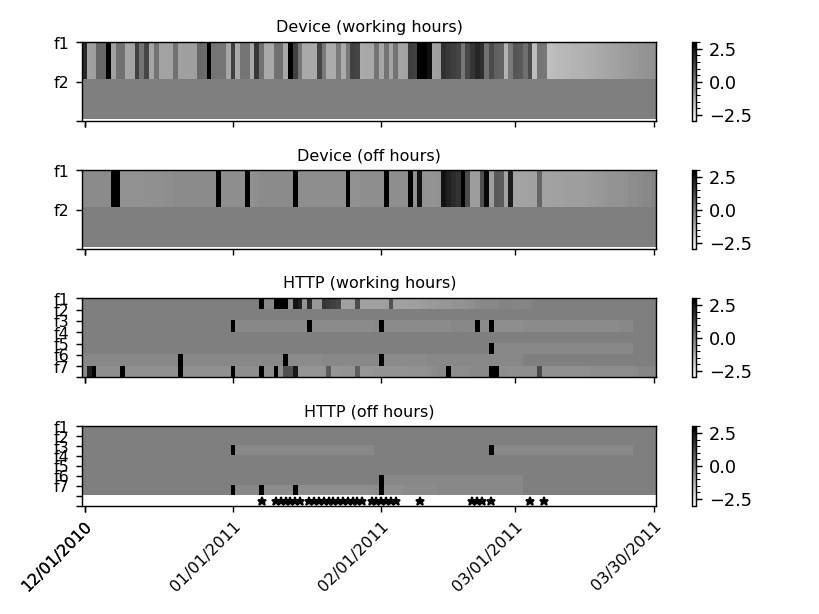}{1}{Example of Abnormal Behavioral Deviations}
\behfig{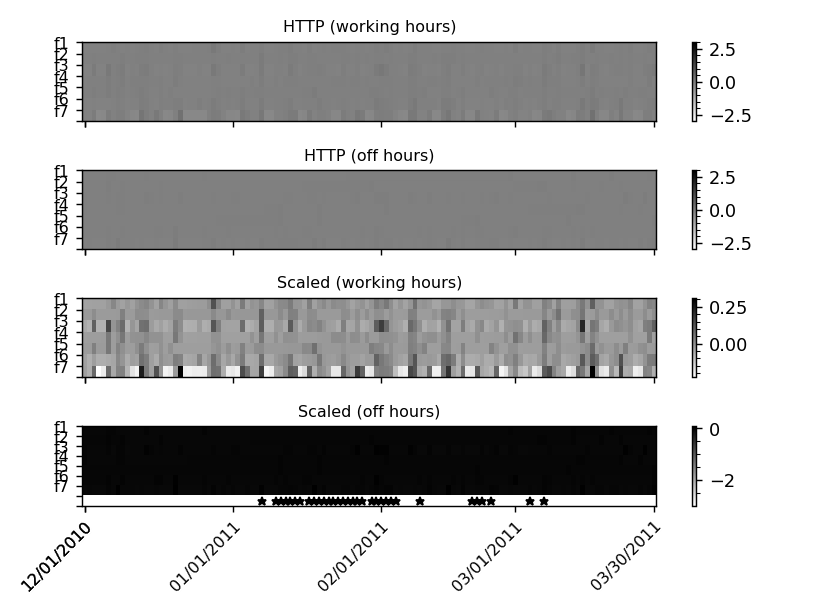}{1}{Example of Group Behavioral Deviations}

The dataset encloses a few types of logs, including device accesses, file accesses, HTTP accesses, email accesses, logon-and-logoffs, and LDAP.  
For presentation purpose, we only present the logs and features that are strongly related to this evaluation. 
For each log type, we split the log entries by user ID, and then for each user ID we extract a set of behavioral deviations $\sigma_{f,t,d}$, each of which represents the number of instances of feature $f$ during the time-frame $t$ on the day $d$.  Feature weights $w_{f,t,d}$ are applied. 
\begin{enumerate}
\item \textbf{\textit{Device Accesses:}}  This category encloses the usage of thumb drives.  Each log entry includes an \textit{activity} (either \textit{connect} or \textit{disconnect}) and a \textit{host ID} to where a thumb drive is connected.  There are in total two deviation features in this category: (f1) \textit{\textbf{connection}}, the number of connections and (f2) \textit{\textbf{new-host-connection}}, the number of connections to a new host that the user never had connected to before day $d$. 

\item \textbf{\textit{File Accesses:}} Each log entry in this category includes an \textit{activity} (e.g., open, copy, write), a \textit{file ID}, and a \textit{dataflow direction}.  There are in total seven features in this category: (f1) \textit{\textbf{open-from-local}}, (f2) \textit{\textbf{open-from-remote}}, (f3) \textit{\textbf{write-to-local}}, (f4) \textit{\textbf{write-to-remote}}, (f5) \textit{\textbf{copy-from-local-to-remote}}, (f6) \textit{\textbf{copy-from-remote-to-local}}, and (f7) \textit{\textbf{new-op}}. The value of each feature is computed as the number of operation in terms of $(\textit{feature}, \textit{file-ID})$ pair that the user never had conducted before day $d$.

\item \textbf{\textit{HTTP Accesses:}} Each log entry in this category includes an \textit{activity} (e.g., visit, download, upload), a \textit{domain}, and a \textit{filetype} that is being downloaded or uploaded.  We do not take \textit{visit} and \textit{download} into consideration, because these two are too chaotic to help with building behavioral profile.  There are in total seven features in this category: (f1) \textit{\textbf{upload-doc}}, (f2) \textit{\textbf{upload-exe}}, (f3) \textit{\textbf{upload-jpg}}, (f4) \textit{\textbf{upload-pdf}}, (f5) \textit{\textbf{upload-txt}}, (f6) \textit{\textbf{upload-zip}}, and
\textit{\textbf{http-new-op}}, (f7). The value of each feature is computed as the number of operation in terms of $(\textit{feature}, \textit{domain})$ pair that the user never had conducted before day $d$).
\end{enumerate}

Figure~\ref{behfig:acobe-r1s2-JPH1910_expbeh.png} depicts the behavioral deviation matrices of the abnormal user \textit{JPH1910}.
The upper two sub-figures are behavioral deviation in the \textit{device-access} aspect (with two features), during working hours and off hours, respectively.  The lower two sub-figures are behavioral deviation in the \textit{HTTP-access} aspect (with seven features), during working hours and off hours, respectively.  The star markers at the bottom indicate the labeled abnormal days; however, unlabeled days are not necessarily normal, as we observed identical events being both labeled and unlabeled.  Behavioral deviations $\sigma_{f,t,d}$ are in range $[-\Delta, \Delta]=[-3,3]$.  We can see in that \textit{JPH1910} has abnormal deviation pattern in the \textit{HTTP upload-doc} (first row) feature starting from January.  These deviations are caused by uploading \textit{``resume.doc''} to several companies, and these events also cause noticeable deviation in the \textit{HTTP new-op} feature (last row).  Dark deviations have white tails, because sliding history window is applied (as they change $mean$ and $std$ that were used in deriving latter $\sigma_{f,t,d}$).  The length of tails hence depends on the window size.  For reference, Figure~\ref{behfig:acobe-r1s2-group_expbeh.png} depicts the group behavioral deviation in the HTTP aspect with two color scales.


\subsection{Research Questions}

\begin{figure*}
\centering    
\subfigure[ACOBE (Device)]{
    \includegraphics[width=0.31\textwidth]
    {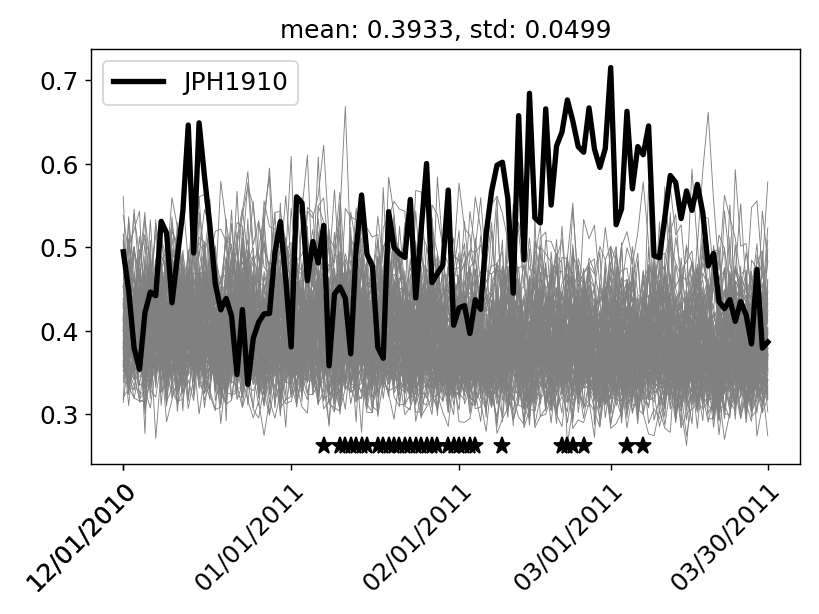}
    \label{behfig:ACOBE:r1s2device.png}} 
\subfigure[ACOBE (HTTP)]{
    \includegraphics[width=0.31\textwidth]
    {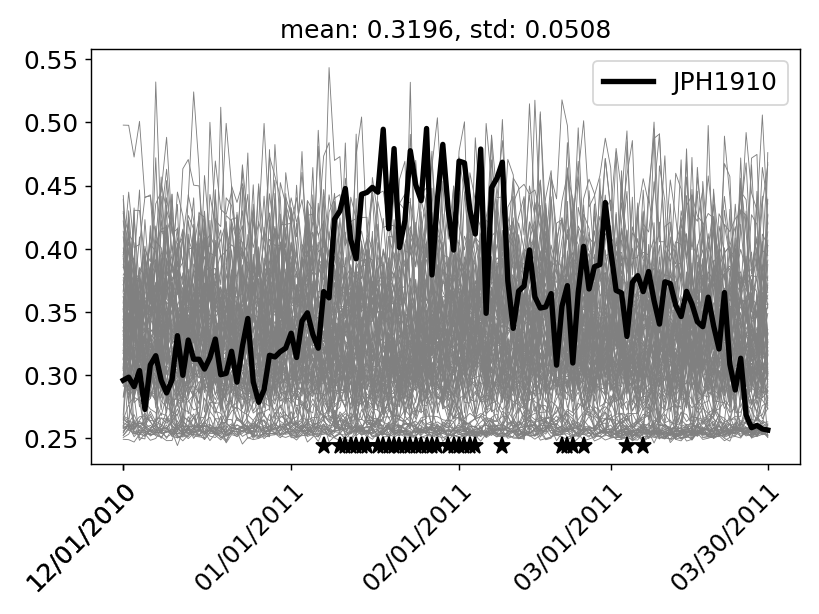}
    \label{behfig:ACOBE:r1s2http.png}}
\subfigure[Single-Day Reconstruction (HTTP)]{
    \includegraphics[width=0.31\textwidth]
    {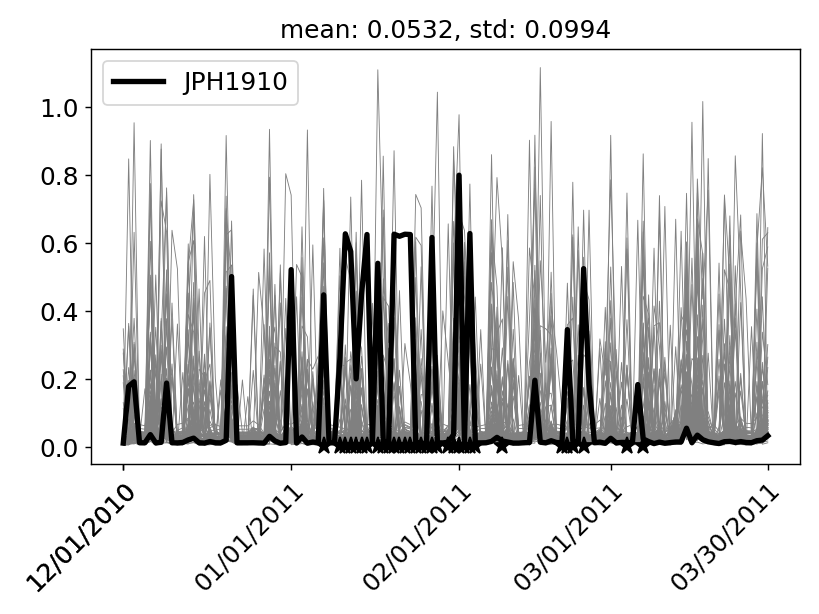}
    \label{behfig:1dr:r1s2http.png}}
    
\subfigure[Excluding Group Deviations (HTTP)]{
    \includegraphics[width=0.31\textwidth]
    {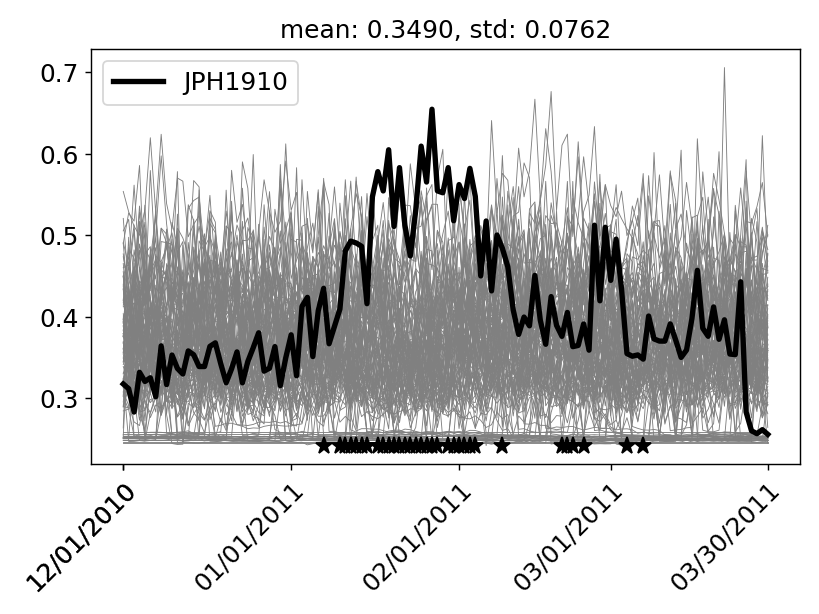}
    \label{behfig:egh:r1s2http.png}} 
\subfigure[All-in-One Autoencoder]{
    \includegraphics[width=0.31\textwidth]
    {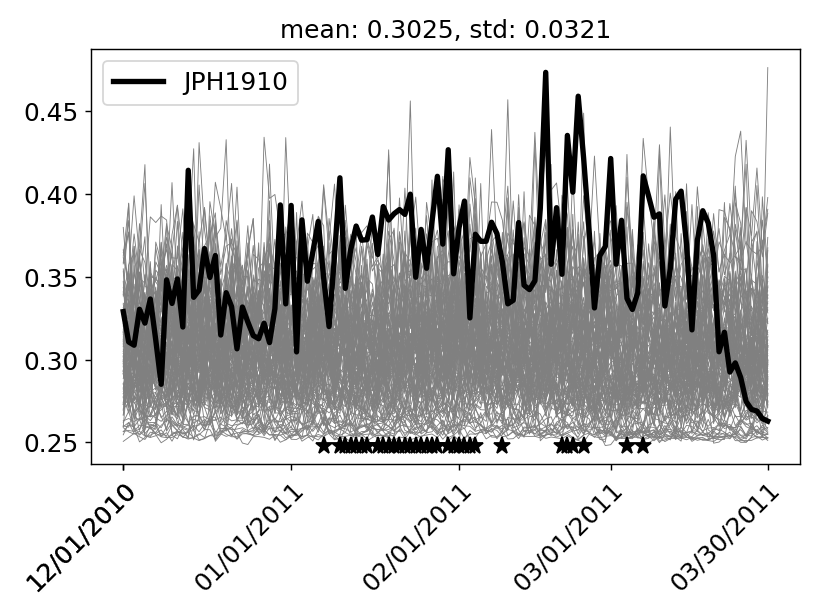}
    \label{behfigure:ain1:r1s2.png}}
\subfigure[Baseline]{
    \includegraphics[width=0.31\textwidth]
    {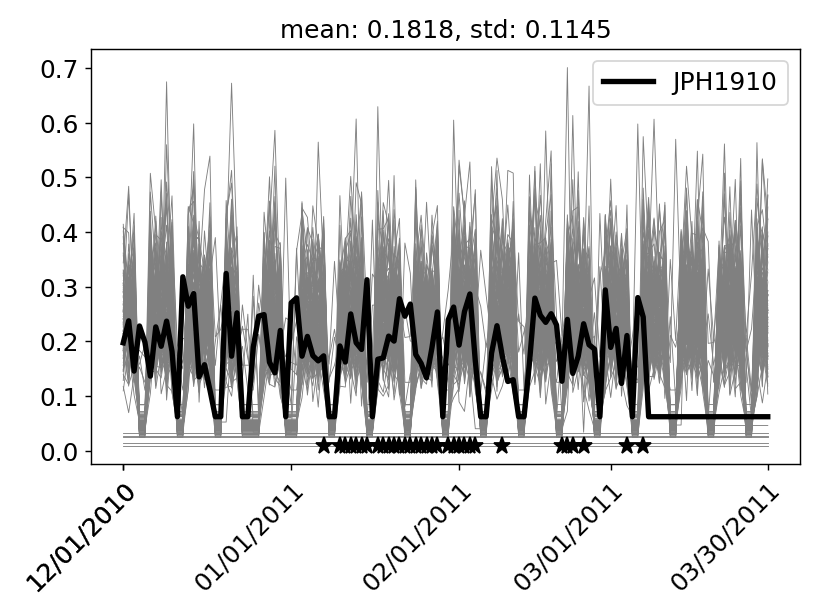}
    \label{behfigure:liu:r1s2.png}}
\caption{Trends of anomaly scores of 114 users in r6.1 Scenario 2 under different model configuration}
\label{behfigure:figureset1}
\end{figure*}

ACOBE has a few different designs compared to prior work (e.g., \cite{liuliu1, pca01}), and we are interested in the following questions:

\begin{enumerate}
\item How does reconstructing multiple days help with detection of abnormal users in a large-scale organization? 
\item How does including group deviations help?
\item How does splitting behavioral aspect help? 
\end{enumerate}

To answer these questions, we present r6.1 Scenario 2 in Figure~\ref{behfigure:figureset1}.  Each sub-figure depicts the trends of anomaly scores of 114 users in the department, to which \textit{JPH1910} belongs.  The black line depicts the score trend of the abnormal user, and the grey lines depict the score trends of 113 normal users.  The star markers at the bottom indicate the labeled abnormal days.  Mean and standard deviation (\textit{std}) are derived by all data points in each sub-figure.   From Figure~\ref{behfig:ACOBE:r1s2device.png} and Figure~\ref{behfig:ACOBE:r1s2http.png}, we can see that \textit{JPH1910} has higher anomaly scores on the dates when we observe the abnormal patterns shown in Figure~\ref{behfig:acobe-r1s2-JPH1910_expbeh.png}.

\subsubsection{\textbf{Long-term vs. Single-Day Reconstruction}}

The reason why we prefer long-term reconstruction rather than single-day reconstruction is that, typical cyber threats (including insider threats) do not start and end within one single day.  Single-day reconstruction cannot identify long-lasting threats that span multiple days.
Figure~\ref{behfig:1dr:r1s2http.png} depicts the anomaly scores derived by a single-day reconstruction model, which is similar to ACOBE except that the features are normalized occurrences of activities (as it no longer has history window for derived behavioral deviations).  
We can see that, although there are abnormal raises on the labeled days, the score waveform of the abnormal user is not distinguishable from the waveforms of normal users.  Thay all have peaks on weekdays and troughs on weekends, and the abnormal user does not have higher anomaly scores on any dates. 
In contrast, as showin in Figure~\ref{behfig:ACOBE:r1s2http.png}, ACOBE with long-term reconstruction can demonstrate the score waveform of the abnormal user, and on some dates the anomaly score stands out on top of all users.
The score gets higher as the abnormal behavior patterns continue to appear in the compound behavioral deviation matrix.  The anomaly scores shortly remain high and then decreases as the abnormal behavioral patterns gradually slides out of the matrix.  
Since the scores remain high as long as the patterns remain in the matrix, the length of plateau depends on the window size, whereas the skew depends on the length of the abnormal patterns.  

\subsubsection{\textbf{With Group Deviations vs. Without Group Deviation}}  
Figure~\ref{behfig:egh:r1s2http.png} depicts the trends of anomaly scores of matrices without group deviations (the other configurations are the same as in ACOBE).  
We can see that although the score waveform of the abnormal user still seems alike and distinguishable as in Figure~\ref{behfig:ACOBE:r1s2http.png}, this model is less ideal, despite that a good model should also be able to identify abnormal users without group deviations (that is, identifying abnormal users only by individual-user behaviors that deviate from the user's history).
A model without considering group deviations is less ideal, because it ignores the behavioral correlation between a user and the group; as a result, this model may overly emphasize self-deviating users, and thus mis-rank normal users before abnormal users (hence false positives).  
Moreover, from Figure~\ref{behfig:egh:r1s2http.png} we can see that the average of anomaly scores is higher (which means reconstruction errors are higher), despite that the size of behavioral matrices is cut in half due to the absence of group deviations.
In contrast, ACOBE's encoder neurons take inputs not only from individual-user deviations but also from group deviations, and they are trained to find the correlations (as in weights and bias) among the inputs.  With such correlations, ACOBE reduces not only the mis-rankings, but also the average of anomaly scores, meaning that group behavior indeed help with reducing reconstruction errors of normal behavioral matrices.
In the following section we show that ACOBE outperforms the corresponding long-term model without group deviations (denoted as \textit{No-Group} model).  

\subsubsection{\textbf{An Ensemble of Autoencoders vs. One Autoencoder}}

The drawback of deploying just one autoencoder for all features is that, the model may be too sensitive to noise introduced by irrelevant features, despite that weights are already applied to features with the assumption that chaotic features are less critical.
ACOBE suffers from the common limitation among all other anomaly detection methods: if a set of features cannot describe a cyber threat, ACOBE may not be able to identify cyber compromises.
Figure~\ref{behfigure:ain1:r1s2.png} depicts the anomaly scores derived by all-in-one model.  The included features are the same features in \textit{device-access}, \textit{file-accesses}, and \textit{HTTP-accesses} aspects; note that, configurations (including the weight function) for this model are identical to the one for ACOBE.  This model is not ideal, because the waveform is not as outstanding as shown in Figure~\ref{behfig:ACOBE:r1s2device.png} considering the usage of thumb drives is critically abnormal under this scenario.  Without emphasis on the \textit{device-accesses} aspect, this model may wrongly report normal users that have slightly chaotic behaviors.
A common approach to resolve the drawback is to deploy an ensemble of autoencoders that each takes responsibility in one aspect (related work includes~\cite{lifelong15, lifelong26, liuliu1, liuliu2}).  This approach gives each autoencoder only necessary features and thus reduces the unwanted noise.  The drawback of deploying an ensemble of autoencoders, however, is the incapability of finding cross-aspect relationship, which is often domain-specific and less interesting. 
To work with multiple autoencoders, ACOBE checks the $N$-th highest rank of each user in every behavioral aspect.  

\subsection{Comparison with Prior Work}

\begin{figure*}
\centering    
\subfigure[ROC (TP Rate - FP Rate) Curve]{
    \includegraphics[width=0.31\textwidth]
    {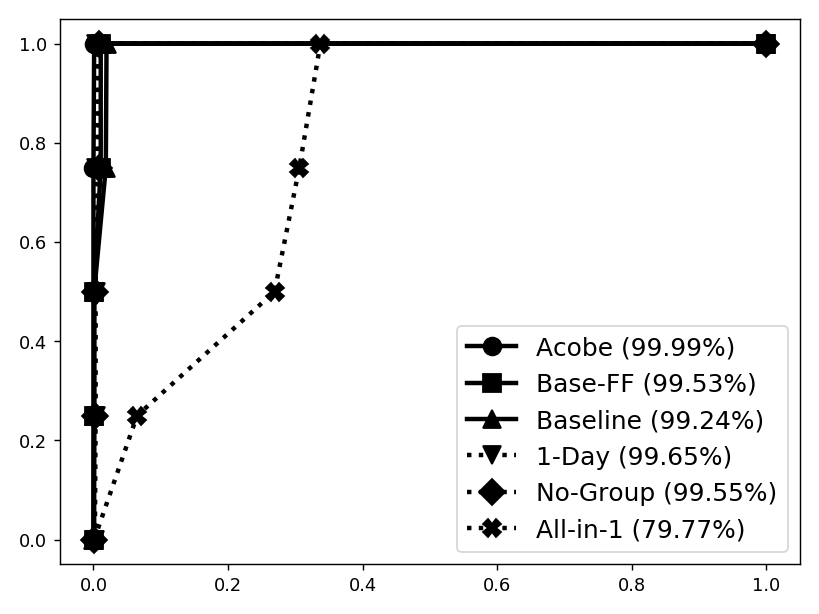}
    \label{behfigure:metric_roc.png}}
\subfigure[Precision-Recall Curve]{
    \includegraphics[width=0.31\textwidth]
    {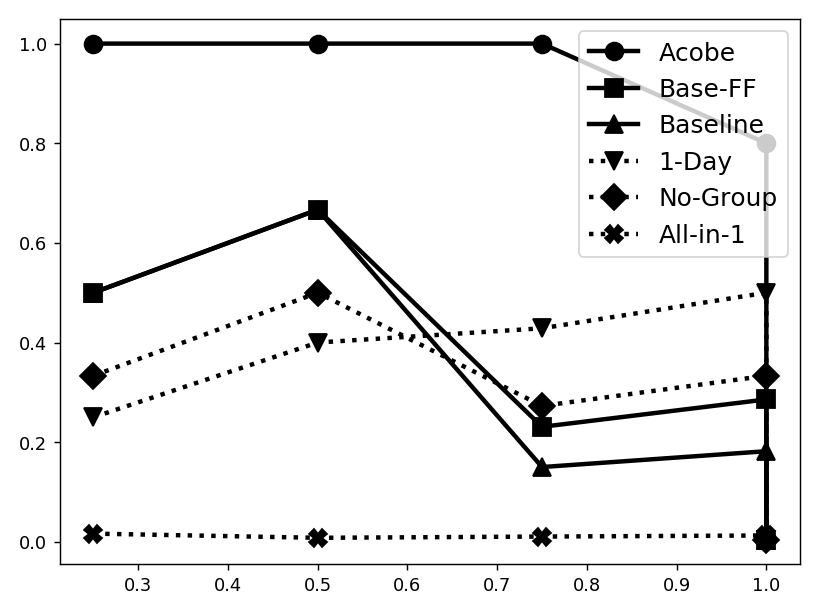}
    \label{behfigure:metric_f1.png}}
\subfigure[Precision-Recall Curve]{
    \includegraphics[width=0.31\textwidth]
    {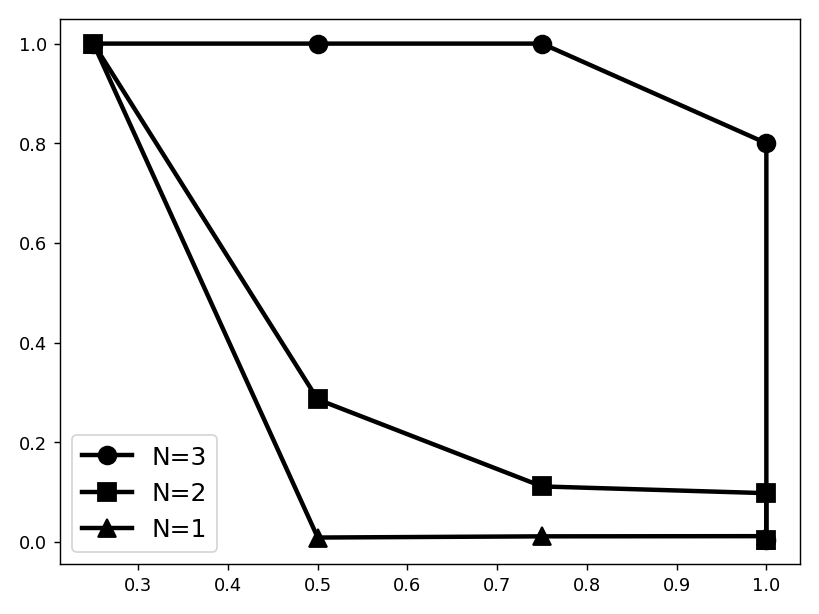}
    \label{behfigure:ACOBE_metric_f1.png}}
\caption{Comparisons Between Models}
\label{behfigure:figureset2}
\end{figure*}

We compare our work with the deep autoencoder model proposed by Liu et al.~\cite{liuliu1}, and we refer to their model to as the \textit{\textbf{Baseline}} model.  The differences between ACOBE and Baseline include the followings.  
First, for each user, Baseline builds four autoencoder for coarse-grained unweighted features from the numbers of \textit{activities} (e.g., connect, write, download, logoff) in four aspects (i.e., \textit{device}, \textit{file} and \textit{HTTP}, and \textit{logon}), whereas ACOBE builds three autoencoders for fine-grained weighted behavioral deviations that also include \textit{new-ops} and \textit{file types}.  
Second, Baseline reconstructs normalized features on individual days, whereas ACOBE reconstructs behavioral deviations across multiple days.  
Third, Baseline does not take group features into consideration, whereas ACOBE embeds group deviations into compound behavioral deviation matrices.  
Fourth, Baseline splits one day into 24 time-frames (i.e., 24 hours), whereas ACOBE splits one day into two (i.e., \textit{working hours} and \textit{off hours}); yet, the number of time-frames contribute negligible performance difference for this dataset.

From the discussions in the previous section, we anticipate that Baseline may not perform well due to the above differences.  Figure~\ref{behfigure:liu:r1s2.png} depicts the anomaly scores of users under r6.1 Scenario 2, and the scores of the abnormal user do not stand out on any dates.  To have a fair comparison, we also build an alternative Baseline model that leverages our fine-grained features, and we refer to the new detection model as the \textit{\textbf{Base-FF}} model.  We implement Baseline and Base-FF with Tensorflow 2.0.0.  Each fully connected layer is implemented with \textit{tensorflow.keras.layers.Dense} activated by ReLU.  The numbers of hidden units at each layer in the encoder are 512, 256, 128, and 64; the numbers of hidden units in the decoder are 64, 128, ,256, and 512.  Between layers, \textit{tensorflow.keras.layers.Batch-Normalization} is applied; it serves the purpose of optimizing training procedure~\cite{batchnormalization}.  To train the model, Adadelta optimizer is used in minimizing the Mean-Squared-Error (MSE) loss function.

We leverage the below metrics to compare the models, where $TP$, $FP$, $TN$, and $FN$ denote the numbers of TPs, FPs, TNs, and FNs, respectively.  These numbers are determined by the number of \textit{how many user investigations the security analysts can conduct}.  For example, say security analysts can investigate $1\%$ of the users, and a user $\mathcal{U}_i$ has its investigation priority ranked at less than $1\%$, then $\mathcal{U}_i$ is considered to be abnormal and will be investigated.  Say, $\mathcal{U}_i$ is however normal, then $\mathcal{U}_i$ is a FP case.  Similarly, say $\mathcal{U}_j$ has its priority ranked greater than $1\%$ and thus is marked as normal while $\mathcal{U}_j$ is indeed abnormal, then $\mathcal{U}_j$ is a FN case.

\begin{align*}
\text{TP Rate} & = \frac{TP}{TP+FN} & \text{FP Rate} & = \frac{FP}{FP+TN} \\ \text{Precision} & = \frac{TP}{TP+FP} & \text{Recall} & = \frac{TP}{TP+FN} 
\end{align*}

Previous work was evaluated by the area under Receiver Operating Characteristic (ROC) curve, where the X-axis represents the FP Rate, and the Y-axis represents the TP Rate. Figure~\ref{behfigure:metric_roc.png} depicts the ROC curves of different models.  The ROC curve is useful when security experts conduct orderly investigations upon ordered users, as the curve depicts the expected trade-offs between TP Rate and FP Rate throughout the investigation process.  Basically, the larger the area under the curve, the better the anomaly detection approach is.  Since we only have four positive (abnormal) cases, we only have four data points on each curve.  Note that, if a FP and a TP has the same top $N$-th rank, the FP is listed before the TP to illustrate the worst-case investigation order. 
For reference purpose, we also include the models we have discussed in previous subsections in this comparison, namely, Excluding Group Deviations (No-Group), Single-Day Reconstruction (1-Day), and All-in-one Autoencoder (All-in-1).

In Figure~\ref{behfigure:metric_roc.png}, the curve of ACOBE is almost a right-angle line, and the areas under ROC curves (AUC) is $99.99\%$.   There are in total 0, and 1 FP (out of 925 normal users) listed before the 3rd and 4th TPs, respectively (that is, the users on top of the investigation list are $[$TP, TP, TP, FP, TP, $\dots]$, and recall that there are only four TPs).  We can see that ACOBE outperforms Baseline (with AUC of $99.23\%$) and Base-FF (with AUC of $99.54\%$), as well as other sub-optimal configurations.  Baseline has 1, 1, 17, and 18 FPs listed before its 1st, 2nd, 3rd, and 4th TPs, respectively.  Base-FF has  1, 1, 10, and 10 FPs, respectively.  If Baseline constantly provides results with $18/925$ FP Rate, security analysts could be overwhelmed in conducting timely investigation depends on the scale of the organization.


However, the ROC metric is known to be misleading for imbalanced dataset~\cite{f1auc}, as it may represent overly optimistic results due to the significantly larger number of negative cases (that is, normal users). We thus also present the Precision-Recall curve in Figure~\ref{behfigure:metric_f1.png}, in which the X-axis represents the recall, and the Y-axis represents the precision.  The importance of this curve includes that the calculations do not make use of the number of TNs, and thus the curve is concerned with only the correct prediction of the small number of positive cases.  
Based on the Precision-Recall curve in Figure~\ref{behfigure:metric_f1.png} we can differentiate ACOBE from Baseline and Base-FF by a large margin.  For reference, while ACOBE's works with $N=3$, we also plot the curves of alternative ACOBE's that work with $N=2$ and $N=1$ in Figure~\ref{behfigure:ACOBE_metric_f1.png}.

Based on the above, we conclude that ACOBE outperforms the Baseline model by a large margin in terms of the Precision-Recall metric.  ACOBE achieves higher area under the Precision-Recall curve, meaning that ACOBE is more effective in correctly ranking abnormal users before normal users.  Hence ACOBE is more accurate in providing an effective investigation list.  

}

    {\section{Case Studies Upon Real Data}\label{behsec:evaluation2}

\begin{figure*}
\centering    
\subfigure[Ransomware's Behavioral Deviation]{
    \includegraphics[width=0.48\textwidth]
    {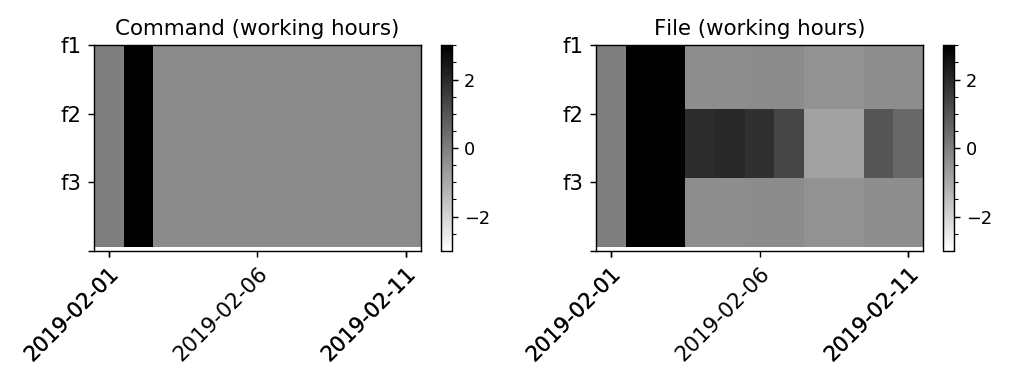}
    \label{behfigure:ransomware_heatmap.png}}
\subfigure[Zeus-Bot's Behavioral Deviation]{
    \includegraphics[width=0.48\textwidth]
    {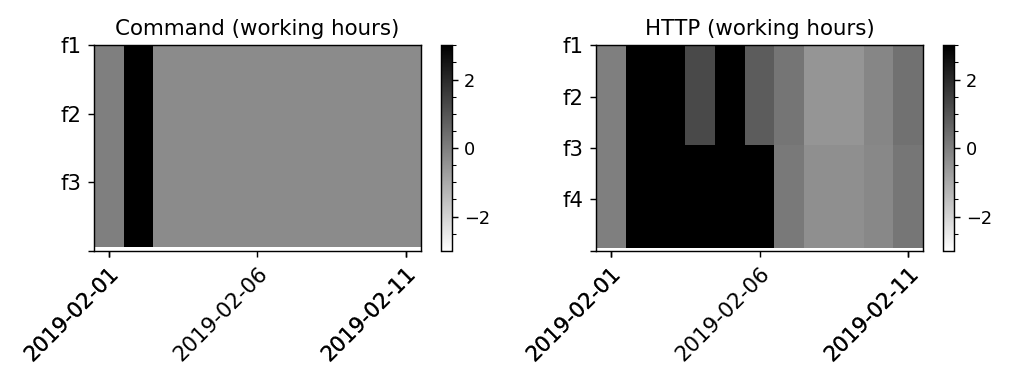}
    \label{behfigure:botnet_heatmap.png}}
\subfigure[Ransomware's Anomaly Score]{
    \includegraphics[width=0.48\textwidth]
    {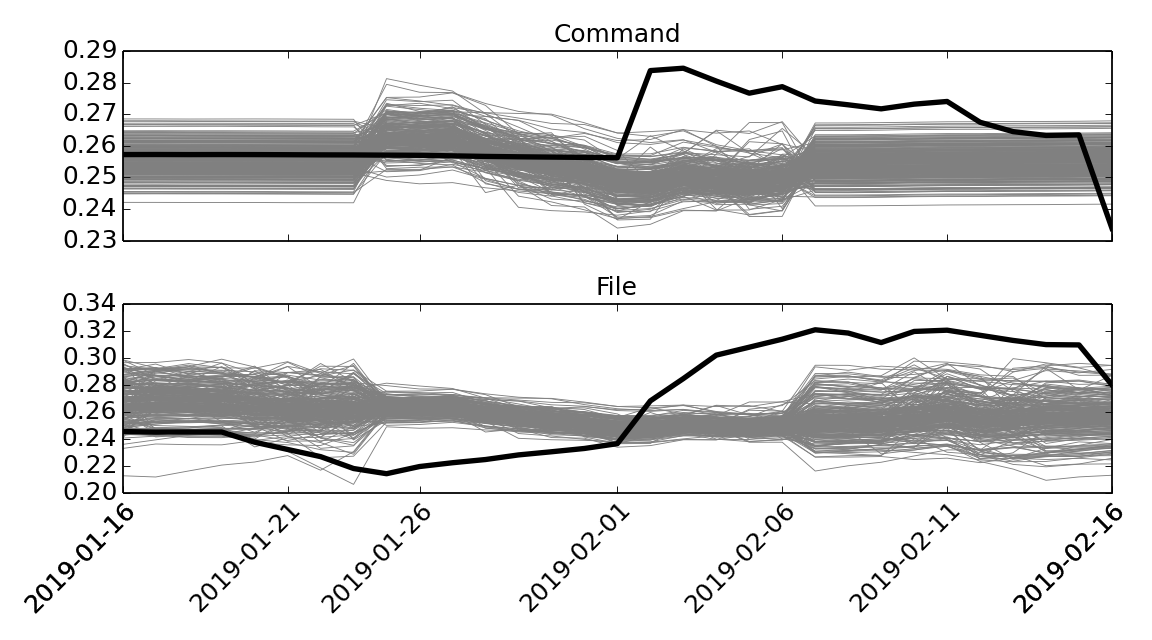}
    \label{behfigure:ransomware_trends.png}}
\subfigure[Zeus-Bot's Anomaly Score]{
    \includegraphics[width=0.48\textwidth]
    {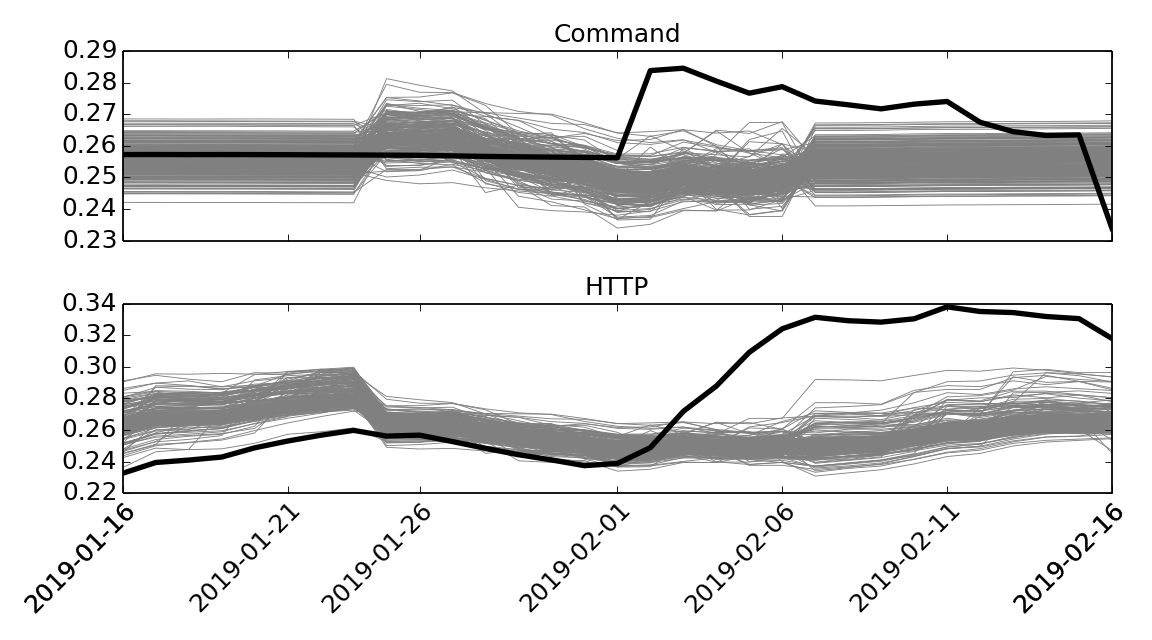}
    \label{behfigure:botnet_trends.png}}
\caption{Case Studies: Ranwomware and Zeus-Bot}
\label{behfigure:figureset3}
\end{figure*}

We applied ACOBE to a real-world enterprise dataset.



%

\subsection{Dataset Engineering}
\textbf{Ethics}: It is important to note that our enterprise set is anonymized where possible privacy concerns are carefully addressed. All identifiable and personal information (e.g., user names/IDs and email addresses) are replaced with securely hashed pseudo information before use.

We gathered a set of audit logs that spans seven months.  Audit logs were generated on Windows servers and web proxies, and they were gathered through the ELK~stack~\cite{elasticsearch}; note that, we do not have audit logs from endpoints (e.g., laptops and mobiles).
Our Windows servers provide logs of the following audit categories: \textit{Windows-Event auditing} (for application, security, setup, and system events), \textit{PowerShell auditing} (\textit{Microsoft-Windows-PowerShell/Operational}), \textit{System-Monitor auditing} (\textit{Microsoft-Windows-Sysmon/Operational}), and DNS-query logs.  To reduce daily log size, we discard events of noisy and redundant event types, including \textit{Process Access} (event ID: 10). 
Web proxies provide its native system and service logs (including syslog and DHCP logs) and informative proxy logs (where each entry includes user, source, destination, resources, and various types of security verdicts).

For presentation purpose, we present only employee accounts (say, \textit{alice} is an employee acount), but employee accounts are integrated with (have activities from) computer accounts (e.g., \textit{alice\$}), email accounts (e.g., \textit{alice@enterprise.com}), and domain accounts (e.g., \textit{ENT\textbackslash alice}, \textit{ENT\textbackslash alice\$}, and \textit{ENT\textbackslash alice@enterprise.com}).  Note that, we exclude service accounts (e.g., \textit{httpd}), and privileged accounts (e.g., admin accounts) from this case study, because they are not real users (e.g., they do not edit Word documents, check emails, or surf web pages).  These accounts are used occasionally only for particularly privileged operations, and thus they do not demonstrate habitual patterns like human. By doing so, we have 246 employees in this dataset.


\nop{
and for the systems, admins, I'll say we are not using them to avoid false labels?
but more importantly is that they are not real users
we often switch to admins and then do something that requires privileges
admin may not have as many operations that does not need priv as normal users
that is, admin does not demonstrate behaviral habitual patterns
but the above is more convincing to me than "avoid false labels"
cuz false label somehow means that we can do admins, it just we have false labels, and we are not fixing false labels.
}

By applying Sigma signatures~\cite{sigma}, we summarize that there were no attacks in the gathered audit logs. 
To work with this dataset, we launched the following two attacks in the same environment under control, and we embed the aftermath attack logs into our enterprise dataset.  
We embed the attack logs into the log of a specific user on specific days, whereas the logs of the other users and the other days are intact. 


\begin{itemize}
\item \textbf{Zeus Botnet:}  Malicious activities include downloading Zeus from a downloader app, deleting this downloader app, and modifying registry values.  Our Zeus bot also made queries to non-existing domains that were generated by \textit{newGOZ} (a domain generation algorithm found in Gameover Zeus and Peer-to-Peer Zeus~\cite{dga, zeus}).  
\item \textbf{Ransomware:}  We use the WannaCry samples available in Malware DB~\cite{zoo}.  Malicious activities include modifying registry values and encrypting files.
\end{itemize}


\subsection{Behavioral Feature Engineering}

We design a set of behavioral features for ACOBE to work with this real-world dataset.
However, note that we want to emphasize only that \textit{``ACOBE can be applied to real audit logs in practice''}, but not that \textit{``we have fairly working features''}. 
It is hard to craft good behavioral features by hand for a complex dataset, and we are aware that our feature design is not perfect.  
For presentation purpose, we split our features into the following two categories; for simplicity, we only present the features that are strongly related to this case study.

\subsubsection{\textbf{Predictable Behavioral Aspects}}

Predictable aspects are aspects where dependency or causality exists among consecutive events, so that we may predict upcoming events based on a sequence of events.  For example, we may predict that a user will read a particular file after gaining a particular file-access permission, or that a user will execute a particular command after one another (e.g., \textit{dir} followed by \textit{ls}).  
To measure how an event sequence deviates from a user's habitual pattern with predictable context, we can leverage deep-learning based anomaly detection model for discrete events (e.g.,DeepLog~\cite{deeplog}).  
In this case study, we present two (out of four) predictable aspects for this dataset: \textit{\textbf{File}} accesses such as file-handle operations, file shares, and Sysmon file-related events (event IDs: 2, 11, 4656, 4658-4663, 4670, 5140-5145), and \textit{\textbf{Command}} executions such as process creation and PowerShell execution (1, 4100-4104, 4688).
\nop{
\begin{enumerate}
\item \textbf{\textit{Intranet Access:}}  When an entity queries an internal resource, an intranet flow is established.  We gather relevant audit logs from web proxies (HTTP logs), domain controllers (DNS logs), and resource servers (IIS logs).  
\item \textbf{\textit{File Access:}}  We gather file-access events happened on Windows servers, including but not limited to, handle operations (event ID: 4656, 4658-4663, 4670), file shares (event ID: 5140-5145), and Sysmon file-related events (event ID: 2, 11) when the operation can be linked with a subject account or SID.  
\item \textbf{\textit{System Configuration}}  We gather Windows system configuration events, including but not limited to, registry modification (event ID: 12-14, 4656-4657), sensitive privilege usage (event ID: 4673-4674), service installation (event ID: 4697), task schedule management (event ID: 4698-4702), and user/group management (event ID: 4717-4767, 4780-4781, 4793-4799). 
\item \textbf{\textit{Command Execution:}}  We gather CommandLine and PowerShell execution events, including but not limited to, process create (event ID: 1, 4688) and PowerShell executing pipeline (event ID: 4100-4104).
\end{enumerate}
}
For simplicity, we present only the following three features for the \textit{\textbf{File}} behavioral aspect and for the \textit{\textbf{Command}} behavioral aspect.   
\begin{itemize}
\item f1: the number of events during a period
\item f2: the number of unique events during a period
\item f3: the number of new events during a period
\end{itemize}

\subsubsection{\textbf{Statistical Behavioral Aspects}}

Statistical aspects are aspects where we cannot predict upcoming events, or aspects that have better solutions for extracting deviation than prediction.
To measure behavioral deviation without predictable context, Siadati et. al~\cite{logonpattern} leverage features that describe statistically structural pattern in an large-scale enterprise.  With observation that each employee has structural access from enterprise account to network resources (e.g., workday portal, cloud drive, social media), we present one (out of two) statistical behavioral aspect: \textit{\textbf{HTTP}} traffic.
For simplicity, we present only the following four features.
\begin{itemize}
\item f1: \# of successful requests during a period $\mathcal{P}$
\item f2: \# of successful requests to a new domain during $\mathcal{P}$
\item f3: \# of failure requests during $\mathcal{P}$
\item f4: \# of failure requests to a new domain during $\mathcal{P}$
\end{itemize}

\subsection{Detection Results}

We showcase how malicious activities impact the victim's behavioral matrix and anomaly scores (with $N=3$) in the two cyber attack scenario.
We have 246 employees, and one of which is under the aforementioned two attacks on Feb 2nd.  
We have in total 27 behavioral features, 16 of which from four behavioral aspects (namely, \textit{File}, \textit{Command}, \textit{Config}, and \textit{Resource}) and 11 from statistical aspects (namely \textit{HTTP} and \textit{Logon}).  The window size for our compound behavioral deviation matrix is two weeks (14 days).  This dataset spans seven months, including six months of training set and one month of testing set.

Figure~\ref{behfigure:ransomware_heatmap.png} and Figure~\ref{behfigure:botnet_heatmap.png} depict the victim's partial compound behavioral deviation matrices.  In both attack scenarios, we see positive deviations in the \textit{Command} aspect.  These deviation are caused by the newly observed executions of the malware programs.  
Such executions increase the number of events (i.e., f1 increases), the number of unique events (i.e., f2 increases), and the number of new events (i.e., f3 increases).  
Since the victim barely has any activities in the \textit{Command} aspect, such deviations are significant (hence, they are dark). 
Though not shown in this paper, we see the same positive deviation in the \textit{Config} aspect, as two attacks both modified registry values shortly after being triggered.
In the ransomware scenario, deviations in the \textit{File} aspect are caused by newly observed \textit{read}, \textit{write}, and \textit{delete} operations conducted by the malware (note that, these operations are no longer new after one day).
In the botnet scenario, deviations in the \textit{HTTP} aspect are caused by successful connections to the C\&C server and failure connections to the \textit{newGOZ} domains.

Figure~\ref{behfigure:ransomware_trends.png} and Figure~\ref{behfigure:botnet_trends.png} further show examples of how each aspect affects the anomaly scores; one aspect is considered per each subfigure. We see that the waveforms have significant rises after the attack day (i.e., Feb 2nd).
Considering all aspects together, our victim is ranked at 1st place in ACOBE's investigation list from Feb 3rd to Feb 15th with both ransomware and botnet malware.  Security analysts can easily find the attacks if periodic investigation is enforced.
In addition, we observe the followings: First, normal users together demonstrate a main stream of score trends.  We can see that normal users have rises in \textit{Command} and drops in \textit{HTTP} on Jan 26th due to a environmental change.  This again indicates that it is important to examine behavioral correlation between an individual and its group.
Second, although the attack day is on Feb 2nd, the waveforms of \textit{File} and \textit{HTTP} do not demonstrate immediate and significant rise for our victim.  This indicates that the widely used single-day detection methodology may not be able to identify the attacks.  In contrast, with long-term deviation pattern embedded in behavioral matrices, the waveforms rise after the attack day.  
Based on the above, we argue that long-term signals and group correlations are very effective in signaling abnormal users.

}

    {\section{Discussion and Future Work}\label{behsec:discussion}

\subsection{Concerns for Feature Extraction}

The features we use in autoencoders are the behavioral deviations from individual users and from their groups; however, feature selection is beyond the scope of this paper, as the selection is domain-specific. 
There is no best way to extract features that can cover all domains, and hence Acobe suffers from the common limitation among anomaly detection methods: if a set of behaviors cannot imply a cyber threat, ACOBE may not be able to identify the compromise.
To help analysts better deploy ACOBE, we list two concerns that should be examined beforehand.  
First, \textit{``what behaviors should be included in features for a particular domain?''}  To resolve this common concern in the cybersecurity community, MITRE ATT\&CK~\cite{mitre} provides a very useful table that links cyber threats with attack techniques.  The table enables analysts to, for each cyber threat, lookup relevant behaviors. 
Second, \textit{``how long the window size should be for matrices?''}  The length of window is domain-specific, and it depends on the timeline of a particular cyber threat.  When the window is too short, ACOBE may not capture abnormal behaviors that are intentionally separated apart on the timeline; yet, when the window is too long, high anomaly scores caused by anomalies will last too long to notice other compromises.  Another factor in deciding window size is whether periodical inspection is intact.  Incorporating matrices of different window sizes is one of our future work.

\subsection{A More Flexible Detection Critic}

The anomaly detection critic in ACOBE is simple: it ranks users in each aspect, and then it check the $N$-th highest rank of each user.  
We do not incorporate a fancy critic, as we only want to showcase the fundamental idea: an autoencoder-based anomaly-detection approach based on compound behaviors.  
Nevertheless, our future work includes an advanced detection critic, which considers more factors other than just ranks.  Other factors include, but not limited to, the followings.  
First, from Figure~\ref{behfigure:figureset1}, we can see that the anomaly score raises significantly once abnormal activities happen; hence, one factor can be \textit{``whether the anomaly score has an unusual spike''}.
Second, assuming periodical inspection is intact and hence recent anomalies are more interesting, then another factor can be \textit{``whether the abnormal raise is recent''}.
Third, different anomalies may demonstrate different characteristics on the anomaly score; for example, normal behavior change, such as a user $\mathcal{U}_i$ has just begun working on a new project, could be a bursting raise with long-lasting but smooth decrease, whereas a cyber attack may not show the decrease but chaotic signals, as the malicious behaviors may not be consistent over time.  Hence, another factor can be \textit{``whether the abnormal raise demonstrate a particular waveform''}.  These additional factors could be useful in anomaly detection judgement, and we plan to study waveform characteristics in the future.

}

    {\section{Conclusion}\label{behsec:conclusion}

The fundamental limitation of the widely-used anomaly-detection methodology, which leverages reconstructions of single-day and individual-user behaviors, is that it overlooks the importance of long-term cyber threats and behavioral correlation within a group, and hence it suffers from large number of false positives.  
To this end, we propose a behavioral representation which we refer to as a \textit{compound behavioral deviation matrix}.  It encloses long-term deviations and group deviations.  
Having such matrices, we then propose ACOBE, an autoencder-based anomaly detection for compromises.
Our evaluation and case studies show that, ACOBE not only outperforms prior related work by a large margin in terms of precision and recall, but also is applicable in practice for discovery of realistic cyber threats.}



\bibliographystyle{IEEEtran}
\bibliography{reference}

\begin{thebibliography}{10}
\providecommand{\url}[1]{#1}
\csname url@samestyle\endcsname
\providecommand{\newblock}{\relax}
\providecommand{\bibinfo}[2]{#2}
\providecommand{\BIBentrySTDinterwordspacing}{\spaceskip=0pt\relax}
\providecommand{\BIBentryALTinterwordstretchfactor}{4}
\providecommand{\BIBentryALTinterwordspacing}{\spaceskip=\fontdimen2\font plus
\BIBentryALTinterwordstretchfactor\fontdimen3\font minus
  \fontdimen4\font\relax}
\providecommand{\BIBforeignlanguage}[2]{{%
\expandafter\ifx\csname l@#1\endcsname\relax
\typeout{** WARNING: IEEEtran.bst: No hyphenation pattern has been}%
\typeout{** loaded for the language `#1'. Using the pattern for}%
\typeout{** the default language instead.}%
\else
\language=\csname l@#1\endcsname
\fi
#2}}
\providecommand{\BIBdecl}{\relax}
\BIBdecl

\bibitem{cyberthreat1}
G.~Belani, ``5 cybersecurity threats to be aware of in 2020,''
  https://www.computer.org/publications/tech-news/trends/5-cybersecurity-threats-to-be-aware-of-in-2020.

\bibitem{cyberthreat2}
D.~Rafter, ``Cyberthreat trends: 15 cybersecurity threats for 2020,''
  https://us.norton.com/internetsecurity-emerging-threats-cyberthreat-trends-cybersecurity-threat-review.html.

\bibitem{svddc}
\BIBentryALTinterwordspacing
T.~Kenaza, K.~Bennaceur, and A.~Labed, ``An efficient hybrid svdd/clustering
  approach for anomaly-based intrusion detection,'' in \emph{Proceedings of the
  33rd Annual ACM Symposium on Applied Computing}, ser. SAC ’18.\hskip 1em
  plus 0.5em minus 0.4em\relax New York, NY, USA: Association for Computing
  Machinery, 2018, p. 435–443. [Online]. Available:
  \url{https://doi.org/10.1145/3167132.3167180}
\BIBentrySTDinterwordspacing

\bibitem{nlsalog16}
Z.~{Liu}, T.~{Qin}, X.~{Guan}, H.~{Jiang}, and C.~{Wang}, ``An integrated
  method for anomaly detection from massive system logs,'' \emph{IEEE Access},
  vol.~6, pp. 30\,602--30\,611, 2018.

\bibitem{lifelong26}
Y.~Mirsky, T.~Doitshman, Y.~Elovici, and A.~Shabtai, ``Kitsune: An ensemble of
  autoencoders for online network intrusion detection,'' 2018.

\bibitem{liuliu1}
L.~{Liu}, O.~{De Vel}, C.~{Chen}, J.~{Zhang}, and Y.~{Xiang}, ``Anomaly-based
  insider threat detection using deep autoencoders,'' in \emph{2018 IEEE
  International Conference on Data Mining Workshops (ICDMW)}, Nov 2018, pp.
  39--48.

\bibitem{liuliu2}
L.~Liu, C.~Chen, J.~Zhang, O.~De~Vel, and Y.~Xiang, ``Unsupervised insider
  detection through neural feature learning and model optimisation,'' in
  \emph{Network and System Security}, J.~K. Liu and X.~Huang, Eds.\hskip 1em
  plus 0.5em minus 0.4em\relax Cham: Springer International Publishing, 2019,
  pp. 18--36.

\bibitem{pca01}
Q.~{Hu}, B.~{Tang}, and D.~{Lin}, ``Anomalous user activity detection in
  enterprise multi-source logs,'' in \emph{2017 IEEE International Conference
  on Data Mining Workshops (ICDMW)}, Nov 2017, pp. 797--803.

\bibitem{pca02}
M.~A. Maloof and G.~D. Stephens, ``Elicit: A system for detecting insiders who
  violate need-to-know,'' in \emph{Proceedings of the 10th International
  Conference on Recent Advances in Intrusion Detection}, ser. RAID’07.\hskip
  1em plus 0.5em minus 0.4em\relax Berlin, Heidelberg: Springer-Verlag, 2007,
  p. 146–166.

\bibitem{deepsurvey1}
\BIBentryALTinterwordspacing
A.~Aldweesh, A.~Derhab, and A.~Z. Emam, ``Deep learning approaches for
  anomaly-based intrusion detection systems: A survey, taxonomy, and open
  issues,'' \emph{Knowledge-Based Systems}, vol. 189, p. 105124, 2020.
  [Online]. Available:
  \url{http://www.sciencedirect.com/science/article/pii/S0950705119304897}
\BIBentrySTDinterwordspacing

\bibitem{deepsurvey2}
R.~Chalapathy and S.~Chawla, ``Deep learning for anomaly detection: A survey,''
  \emph{CoRR}, 2019.

\bibitem{deepsurvey3}
\BIBentryALTinterwordspacing
F.~Falc\~{a}o, T.~Zoppi, C.~B.~V. Silva, A.~Santos, B.~Fonseca, A.~Ceccarelli,
  and A.~Bondavalli, ``Quantitative comparison of unsupervised anomaly
  detection algorithms for intrusion detection,'' in \emph{Proceedings of the
  34th ACM/SIGAPP Symposium on Applied Computing}, ser. SAC ’19.\hskip 1em
  plus 0.5em minus 0.4em\relax New York, NY, USA: Association for Computing
  Machinery, 2019, p. 318–327. [Online]. Available:
  \url{https://doi.org/10.1145/3297280.3297314}
\BIBentrySTDinterwordspacing

\bibitem{dataquality}
A.~{Sundararajan}, T.~{Khan}, A.~{Moghadasi}, and A.~I. {Sarwat}, ``Survey on
  synchrophasor data quality and cybersecurity challenges, and evaluation of
  their interdependencies,'' \emph{Journal of Modern Power Systems and Clean
  Energy}, vol.~7, no.~3, pp. 449--467, 2019.

\bibitem{certdataset2}
J.~{Glasser} and B.~{Lindauer}, ``Bridging the gap: A pragmatic approach to
  generating insider threat data,'' in \emph{2013 IEEE Security and Privacy
  Workshops}, 2013, pp. 98--104.

\bibitem{certdataset1}
``Insider threat test dataset,''
  https://resources.sei.cmu.edu/library/asset-view.cfm?assetid=508099.

\bibitem{log2vec}
F.~Liu, Y.~Wen, D.~Zhang, X.~Jiang, X.~Xing, and D.~Meng, ``Log2vec: A
  heterogeneous graph embedding based approach for detecting cyber threats
  within enterprise,'' in \emph{Proceedings of the 2019 ACM SIGSAC Conference
  on Computer and Communications Security}, ser. CCS ’19.\hskip 1em plus
  0.5em minus 0.4em\relax New York, NY, USA: Association for Computing
  Machinery, 2019, p. 1777–1794.

\bibitem{nlsalog20}
A.~{Oprea}, Z.~{Li}, T.~{Yen}, S.~H. {Chin}, and S.~{Alrwais}, ``Detection of
  early-stage enterprise infection by mining large-scale log data,'' in
  \emph{2015 45th Annual IEEE/IFIP International Conference on Dependable
  Systems and Networks}, June 2015, pp. 45--56.

\bibitem{lifelong}
\BIBentryALTinterwordspacing
M.~Du, Z.~Chen, C.~Liu, R.~Oak, and D.~Song, ``Lifelong anomaly detection
  through unlearning,'' in \emph{Proceedings of the 2019 ACM SIGSAC Conference
  on Computer and Communications Security}, ser. CCS ’19.\hskip 1em plus
  0.5em minus 0.4em\relax New York, NY, USA: Association for Computing
  Machinery, 2019, p. 1283–1297. [Online]. Available:
  \url{https://doi.org/10.1145/3319535.3363226}
\BIBentrySTDinterwordspacing

\bibitem{lifelong48}
\BIBentryALTinterwordspacing
B.~Zong, Q.~Song, M.~R. Min, W.~Cheng, C.~Lumezanu, D.~Cho, and H.~Chen, ``Deep
  autoencoding gaussian mixture model for unsupervised anomaly detection,'' in
  \emph{International Conference on Learning Representations}, 2018. [Online].
  Available: \url{https://openreview.net/forum?id=BJJLHbb0-}
\BIBentrySTDinterwordspacing

\bibitem{chiba}
\BIBentryALTinterwordspacing
Z.~Chiba, N.~Abghour, K.~Moussaid, A.~E. Omri, and M.~Rida, ``A novel
  architecture combined with optimal parameters for back propagation neural
  networks applied to anomaly network intrusion detection,'' \emph{Computers \&
  Security}, vol.~75, pp. 36 -- 58, 2018. [Online]. Available:
  \url{http://www.sciencedirect.com/science/article/pii/S0167404818300543}
\BIBentrySTDinterwordspacing

\bibitem{lifelong31}
\BIBentryALTinterwordspacing
M.~Sakurada and T.~Yairi, ``Anomaly detection using autoencoders with nonlinear
  dimensionality reduction,'' in \emph{Proceedings of the MLSDA 2014 2nd
  Workshop on Machine Learning for Sensory Data Analysis}, ser.
  MLSDA’14.\hskip 1em plus 0.5em minus 0.4em\relax New York, NY, USA:
  Association for Computing Machinery, 2014, p. 4–11. [Online]. Available:
  \url{https://doi.org/10.1145/2689746.2689747}
\BIBentrySTDinterwordspacing

\bibitem{exploitembed}
X.~{Lu}, W.~{Zhang}, and J.~{Huang}, ``Exploiting embedding manifold of
  autoencoders for hyperspectral anomaly detection,'' \emph{IEEE Transactions
  on Geoscience and Remote Sensing}, vol.~58, no.~3, pp. 1527--1537, March
  2020.

\bibitem{gee}
Q.~P. {Nguyen}, K.~W. {Lim}, D.~M. {Divakaran}, K.~H. {Low}, and M.~C. {Chan},
  ``Gee: A gradient-based explainable variational autoencoder for network
  anomaly detection,'' in \emph{2019 IEEE Conference on Communications and
  Network Security (CNS)}, June 2019, pp. 91--99.

\bibitem{advae}
\BIBentryALTinterwordspacing
X.~Wang, Y.~Du, S.~Lin, P.~Cui, Y.~Shen, and Y.~Yang, ``advae: A
  self-adversarial variational autoencoder with gaussian anomaly prior
  knowledge for anomaly detection,'' \emph{Knowledge-Based Systems}, vol. 190,
  p. 105187, 2020. [Online]. Available:
  \url{http://www.sciencedirect.com/science/article/pii/S0950705119305283}
\BIBentrySTDinterwordspacing

\bibitem{lifelong15}
M.~Alam, J.~Gottschlich, N.~Tatbul, J.~Turek, T.~Mattson, and A.~Muzahid, ``A
  zero-positive learning approach for diagnosing software performance
  regressions,'' 2017.

\bibitem{rcae}
R.~Chalapathy, A.~K. Menon, and S.~Chawla, ``Robust, deep and inductive anomaly
  detection,'' in \emph{Machine Learning and Knowledge Discovery in Databases},
  M.~Ceci, J.~Hollm{\'e}n, L.~Todorovski, C.~Vens, and S.~D{\v{z}}eroski,
  Eds.\hskip 1em plus 0.5em minus 0.4em\relax Cham: Springer International
  Publishing, 2017, pp. 36--51.

\bibitem{lifelong47}
\BIBentryALTinterwordspacing
C.~Zhou and R.~C. Paffenroth, ``Anomaly detection with robust deep
  autoencoders,'' in \emph{Proceedings of the 23rd ACM SIGKDD International
  Conference on Knowledge Discovery and Data Mining}, ser. KDD ’17.\hskip 1em
  plus 0.5em minus 0.4em\relax New York, NY, USA: Association for Computing
  Machinery, 2017, p. 665–674. [Online]. Available:
  \url{https://doi.org/10.1145/3097983.3098052}
\BIBentrySTDinterwordspacing

\bibitem{deepsurvey4}
V.~{Chandola}, A.~{Banerjee}, and V.~{Kumar}, ``Anomaly detection for discrete
  sequences: A survey,'' \emph{IEEE Transactions on Knowledge and Data
  Engineering}, vol.~24, no.~5, pp. 823--839, May 2012.

\bibitem{survey1}
A.~L. {Buczak} and E.~{Guven}, ``A survey of data mining and machine learning
  methods for cyber security intrusion detection,'' \emph{IEEE Communications
  Surveys Tutorials}, vol.~18, no.~2, pp. 1153--1176, Secondquarter 2016.

\bibitem{nlsalog10}
\BIBentryALTinterwordspacing
Q.~Fu, J.-G. Lou, Y.~Wang, and J.~Li, ``Execution anomaly detection in
  distributed systems through unstructured log analysis,'' in
  \emph{International conference on Data Mining (full paper)}.\hskip 1em plus
  0.5em minus 0.4em\relax IEEE, December 2009. [Online]. Available:
  \url{https://www.microsoft.com/en-us/research/publication/execution-anomaly-detection-in-distributed-systems-through-unstructured-log-analysis/}
\BIBentrySTDinterwordspacing

\bibitem{nlsalog25}
S.~{He}, J.~{Zhu}, P.~{He}, and M.~R. {Lyu}, ``Experience report: System log
  analysis for anomaly detection,'' in \emph{2016 IEEE 27th International
  Symposium on Software Reliability Engineering (ISSRE)}, Oct 2016, pp.
  207--218.

\bibitem{oprea2018made}
A.~Oprea, Z.~Li, R.~Norris, and K.~Bowers, ``Made: Security analytics for
  enterprise threat detection,'' in \emph{Proceedings of the 34th Annual
  Computer Security Applications Conference}, 2018, pp. 124--136.

\bibitem{stackautoencoder}
P.~Vincent, H.~Larochelle, I.~Lajoie, Y.~Bengio, and P.-A. Manzagol, ``Stacked
  denoising autoencoders: Learning useful representations in a deep network
  with a local denoising criterion,'' \emph{J. Mach. Learn. Res.}, vol.~11, p.
  3371–3408, Dec. 2010.

\bibitem{batchnormalization}
S.~Ioffe and C.~Szegedy, ``Batch normalization: Accelerating deep network
  training by reducing internal covariate shift,'' 2015.

\bibitem{f1auc}
\BIBentryALTinterwordspacing
T.~Saito and M.~Rehmsmeier, ``The precision-recall plot is more informative
  than the roc plot when evaluating binary classifiers on imbalanced
  datasets,'' \emph{PLOS ONE}, vol.~10, no.~3, pp. 1--21, 03 2015. [Online].
  Available: \url{https://doi.org/10.1371/journal.pone.0118432}
\BIBentrySTDinterwordspacing

\bibitem{elasticsearch}
``Elasticsearch: Restful, distributed search and analytics,''
  https://www.elastic.co/.

\bibitem{sigma}
``Sigma: Generic signature format for siem systems,''
  https://github.com/Neo23x0/sigma.

\bibitem{dga}
J.~Bacher, ``Domain generation algorithms,''
  \\https://github.com/baderj/domain\_generation\_\\algorithms.

\bibitem{zeus}
``Zeus,'' https://github.com/Visgean/Zeus.

\bibitem{zoo}
``thezoo aka malware db - a live malware repository,''
  \\https://thezoo.morirt.com/.

\bibitem{deeplog}
M.~Du, F.~Li, G.~Zheng, and V.~Srikumar, ``Deeplog: Anomaly detection and
  diagnosis from system logs through deep learning,'' in \emph{Proceedings of
  the 2017 ACM SIGSAC Conference on Computer and Communications Security}, ser.
  CCS '17.\hskip 1em plus 0.5em minus 0.4em\relax New York, NY, USA: ACM, 2017,
  pp. 1285--1298.

\bibitem{logonpattern}
\BIBentryALTinterwordspacing
H.~Siadati and N.~Memon, ``Detecting structurally anomalous logins within
  enterprise networks,'' in \emph{Proceedings of the 2017 ACM SIGSAC Conference
  on Computer and Communications Security}, ser. CCS '17.\hskip 1em plus 0.5em
  minus 0.4em\relax New York, NY, USA: ACM, 2017, pp. 1273--1284. [Online].
  Available: \url{http://doi.acm.org/10.1145/3133956.3134003}
\BIBentrySTDinterwordspacing

\bibitem{mitre}
``Mitre att\&ck,'' https://attack.mitre.org/.

\end{thebibliography}

\end{document}